\documentclass[letterpaper, 10 pt, conference]{ieeeconf}  

\usepackage{graphicx}
\usepackage{amsmath}
\usepackage{amssymb}
\usepackage{lipsum} 
\usepackage{placeins}
\usepackage{subcaption} 
\usepackage{caption}
\usepackage{xcolor}
\usepackage{hyperref}
\hypersetup{
    colorlinks = true,
    linkcolor = blue,
    urlcolor = blue,
}

\usepackage{booktabs}
\usepackage{cite}
\let\labelindent\relax
\usepackage{enumitem}
\usepackage{multirow}
\usepackage{tabularx}
\usepackage{algorithm}
\usepackage{algorithmicx}
\usepackage{algpseudocode}
\usepackage{url}
\usepackage{siunitx}

\newcommand{\plus}{\thinspace\raisebox{0pt}[0pt][0pt]{+}\thinspace}
\newcommand{\model}{CARL} 
\newcommand{\nop}[1]{}

\IEEEoverridecommandlockouts                              
\title{\LARGE \bf
\model: Congestion-Aware Reinforcement Learning for Imitation-based Perturbations in Mixed Traffic Control 
}
\author{Bibek Poudel$^{1}$, Weizi Li$^{1}$, and Shuai Li$^{2}$ 
\thanks{$^{1}$Bibek Poudel and Weizi Li are with Min H. Kao Department of Electrical Engineering and Computer Science at University of Tennessee, Knoxville, TN, USA {\tt\small bpoudel3@vols.utk.edu, weizili@utk.edu}}%
\thanks{$^{2}$Shuai Li is with Department of Civil and Environmental Engineering at University of Tennessee, Knoxville, TN, USA {\tt\small sli48@utk.edu}}
}

\begin{document}

\maketitle
\thispagestyle{empty}
\pagestyle{empty}


\begin{abstract}
Human-driven vehicles (HVs) exhibit complex and diverse behaviors. Accurately modeling such behavior is crucial for validating Robot Vehicles (RVs) in simulation and realizing the potential of mixed traffic control. However, existing approaches like parameterized models and data-driven techniques struggle to capture the full complexity and diversity. To address this, in this work, we introduce \model{}, a hybrid approach that combines imitation learning for close proximity car-following and probabilistic sampling for larger headways. We also propose two classes of RL-based RVs: a safety RV focused on maximizing safety and an efficiency RV focused on maximizing efficiency. Our experiments show that the safety RV increases Time-to-Collision above the critical $4$ second threshold and reduces Deceleration Rate to Avoid a Crash by up to $80\%$, while the efficiency RV achieves improvements in throughput of up to $49\%$. These results demonstrate the effectiveness of \model{} in enhancing both safety and efficiency in mixed traffic.

\end{abstract}

\section{Introduction}

Robot Vehicles (RVs) have the potential to revolutionize transportation by enhancing safety, efficiency, and accessibility for all road users~\cite{usdot_avcp}. 
To fully realize the benefits of this technology, it is crucial to validate their robustness through comprehensive testing~\cite{epa_selfdriving}. While real-world evaluation is essential, it involves significant risks, costs, and time constraints, making simulation a safer, more efficient, and cost-effective alternative~\cite{bhattacharyyamulti}.
Nevertheless, the validation of RVs in simulation faces the challenge of accurately modeling the behaviors of Human-driven Vehicles (HVs)~\cite{kuefler2017imitating}.
As more vehicles with varying levels of autonomy are introduced into our transportation system, the idea of \textit{mixed traffic control}, which involves the use of RVs to mitigate problems such as congestion and delays produced by HVs, has emerged~\cite{di2021survey,Villarreal2023Pixel,Villarreal2024Eco,Villarreal2023Can,Wang2023Intersection,Wang2024Privacy,Poudel2024EnduRL}. In mixed traffic scenarios where RVs and HVs co-exist, accurately modeling real-world human driving behavior becomes even more critical. However, this remains an open problem~\cite{hawke2020urban}.

\begin{figure}[t!]
		\centering
        \vspace{2pt}
		\includegraphics[width=0.85\linewidth]{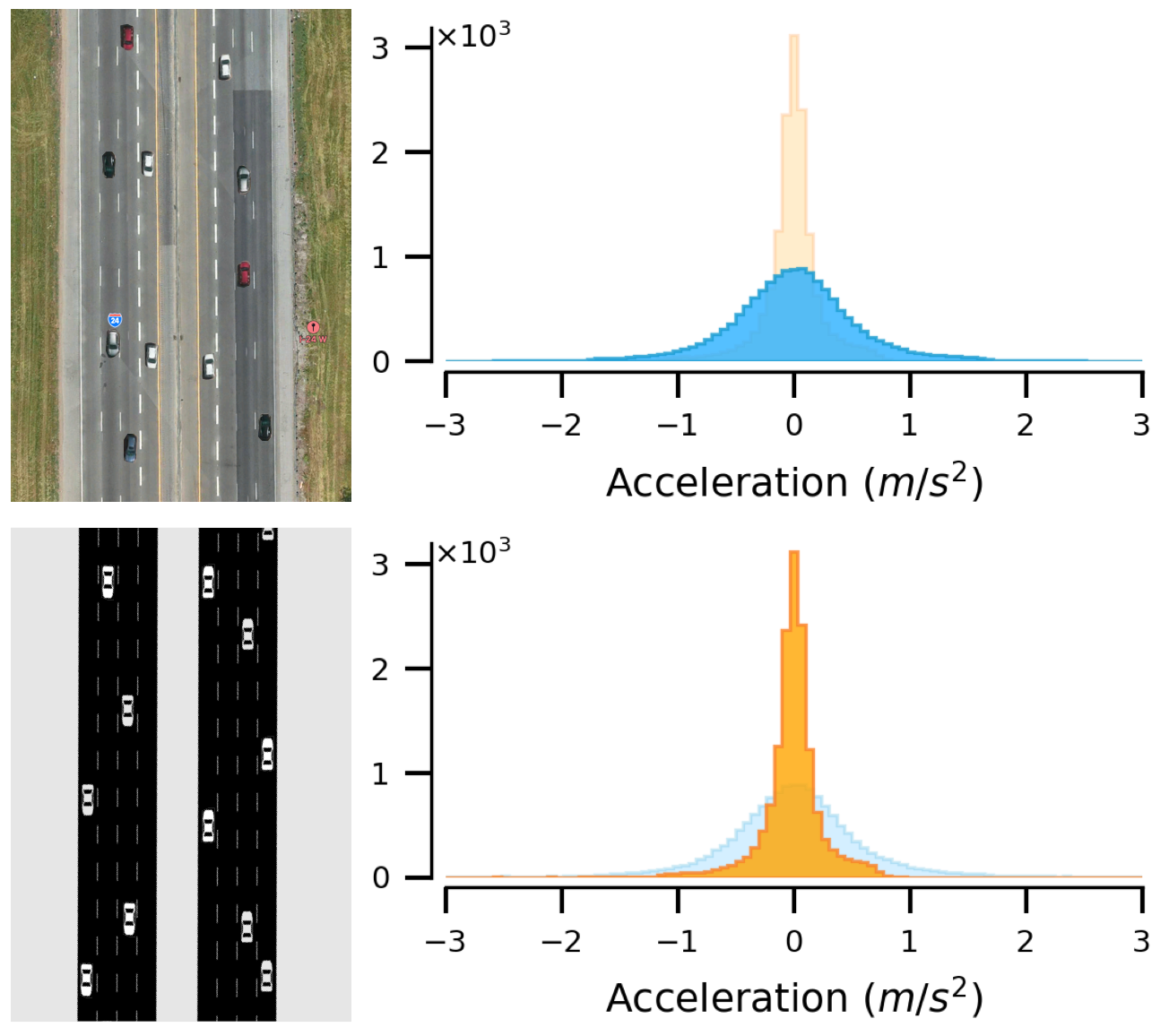}
        \vspace{1pt}
		\caption{\small{Instantaneous accelerations observed during car-following behaviors at densities $[70, 150]~veh/km$. TOP: Real-world data from the I-$24$ MOTION dataset reveals a distribution having long tails extending to $[-3, 3]~m/s^2$. BOTTOM: IDM (in simulation) produces accelerations mostly within $[-0.5, 0.5]~m/s^2$, indicating much `timid' driving behaviors than the real world.}}
		\label{fig:real_world}
    \vspace{-16pt}
\end{figure}

Among the various aspects of real-world human driving, longitudinal car-following is the most prevalent~\cite{zhang2021comprehensive}.
Consequently, accurate modeling of car-following behavior is crucial for reproducing realistic traffic flows and vehicle interactions in simulation, and is a key component in addressing the \textit{sim-$2$-real} gap~\cite{lichtle2022deploying}. To reproduce the car-following behaviors of HVs in simulation, there exist two mainstream methods: parameterized models and data-driven approaches. Popular parameterized car-following models, such as Gipp's~\cite{gipps1981behavioural}, Krauss'\cite{krauss1998microscopic}, and the Intelligent Driver Model (IDM)\cite{treiber2013traffic}, rely on calibrating various parameters (typically include maximum acceleration, minimum time gap, and desired velocity) to accurately represent driving behaviors and vehicle dynamics. However, the performance of these models heavily depends on the quality of calibration, and even with robust calibration, they may still fall short in capturing traffic diversity and lack broad applicability~\cite{albeaik2022limitations}. Enhancements to the IDM model (often a default choice in simulations like SUMO~\cite{lopez2018microscopic}) by addition of random noise~\cite{treiber2017intelligent} and tuning with real-world data~\cite{kesting2008calibrating,li2016global} have been proposed, but they struggle to reproduce real-world variability~\cite{sharath2020enhanced} and fail to generalize outside the reference datasets~\cite{zhu2018modeling}. Fig.~\ref{fig:real_world} illustrates a typical example, where the IDM with random noise model fails to capture the long-tailed distribution of accelerations observed in real-world data, exhibiting a $24\%$ discrepancy (between real-world and simulation) in the percentage of accelerations observed within $[-0.5, 0.5]~m/s^2$, i.e., depicting primarily safe or timid behaviors in simulation. Further, prior studies in traffic control and coordination often impose artificial limits on vehicle behaviors, such as bounded accelerations~\cite{wu2017flow, kreidieh2018dissipating, chou2022lord, vinitsky2018benchmarks, sridhar2021piecewise}, which further reduce the accuracy of simulated HVs.

Data-driven approaches such as supervised learning and Reinforcement Learning (RL) have surpassed IDM in car-following simulation accuracy~\cite{wang2017capturing}. These techniques train neural networks on features extracted from real-world data~\cite{wang2017capturing} or use the data to tune the reward function in RL~\cite{zhu2020safe, zhu2018human}. Some RL-based approaches also equip HVs with both leader and follower vehicle information~\cite{shi2022bilateral}. However, these methods still face challenges in capturing the full complexity of human driving, as they often require hand-crafted features and extensive training and tuning~\cite{bhattacharyya2022modeling}. Imitation Learning (IL) offers a promising alternative to address these shortcomings. By learning directly from expert demonstrations, IL captures the implicit knowledge and individual preferences of human drivers, such as safety and comfort, without the need to specify these objectives. In addition, IL models can be more robust under environmental uncertainties and disturbances, as the expert's behavior may already account for these factors~\cite{zhang2023review}. While IL shows great potential, relying solely on it may still not be sufficient to capture the full spectrum of human driving behaviors and reactions as IL may overfit and struggle to generalize outside the training distribution~\cite{le2022survey}. 

To address these limitations, we introduce \model{}, a hybrid technique that combines imitation learning and probabilistic sampling. The key difference in our work is that we consider the proximity between HVs during car-following to be a crucial factor in determining the appropriate acceleration model. When HVs are at close proximity to a leader HV during car-following, their accelerations are obtained from the imitation learning model, which captures the nuances of human driving behavior in these more sensitive situations, as the consequences of small changes in accelerations are more profound and potentially dangerous. Conversely, when the space headway exceeds a threshold, we employ probabilistic sampling to generate accelerations that introduce realistic variability. This hybrid approach enables \model{} to produce more accurate representations of real-world driving behaviors, including aggressive acceleration profiles. \model{} leverages the strengths of both machine learning methods, which offer accurate representations of real-world driving behaviors, and probabilistic sampling, which minimizes approximations and assumptions. In addition, we propose two classes of RL-based RVs: a \textit{safety} RV focused on maximizing safety and an \textit{efficiency} RV focused on maximizing efficiency. Our RVs are designed to optimize their actions based on the congestion conditions predicted by a supervised classifier, whose output is incorporated into the observation and reward of the RL algorithm. 

Under the realistic-accelerations we evaluate the safety and efficiency and compare \model{} with RVs proposed in prior studies for mixed traffic control. To evaluate safety, we use two surrogate measures, time to collision (TTC $\uparrow$) and deceleration rate to avoid a crash (DRAC $\downarrow$) whereas for efficiency we measure fuel economy (FE $\uparrow$) and throughput ($\uparrow$). Our experiments show that our RV consistently outperforms other methods in safety, increasing TTC above $4~s$ and reducing DRAC by up to $80\%$. Whereas in efficiency our RV achieves improvements of up to $49\%$ in throughput while also consistently maintaining the second-highest fuel economy among all evaluated methods. These results show that \model{} effectively improves both safety and efficiency in mixed traffic. To the best of our knowledge, \model{} is the first work to address the crucial gap between simulated and real-world car-following behaviors using a hybrid imitation learning approach, and to leverage congestion-aware RL for optimizing mixed traffic control. The project code can be found in the repository: \underline{\url{https://github.com/poudel-bibek/CARL}}.

\section{Methodology}
\label{sec:methodology}


We introduce our data processing procedure, Intelligent Driver Model (IDM), Model-based and Heuristic-based RVs, our RL-based RV, and the imitation of real-world driving behaviors and perturbations.  

\subsection{Data Processing and Intelligent Driver Model (IDM)} 
\label{subsec:filter}
We apply a car-following-filter~\cite{zhu2018modeling} to the I-$24$ MOTION dataset~\cite{gloudemans202324} with $6.75~km$ study length and $4~h$ study time. The dataset contains different vehicle types such as semi-trailers, mid-sized trucks, motorbikes, and cars under various traffic conditions such as approaching standing traffic, lane changing, and free flow. 
To extract car-following trajectories, we select data points that meet the following criteria:

\begin{itemize} 
    \item Ego car is following another car, i.e., has a leader.
    \item Leader and ego cars are in the same lane $\ge 5~s$.
    \item Ego car's speed is $>10\%$ of the speed limit, i.e., not approaching stationary traffic. 
    \item Ego car's space headway is $<124~m$, applying $4~s$ rule at the speed limit to avoid free flow conditions.
\end{itemize}

\noindent By applying the filter, we extract a total of $172,000$ instantaneous accelerations, their distribution is shown in Fig.\ref{fig:real_world} TOP. In comparision, Fig.~\ref{fig:real_world} BOTTOM depicts the accelerations obtained from the IDM car-following model, which assumes that drivers strive to maintain a safe distance from the leader vehicle while trying to achieve a target speed. In IDM, vehicles speed up when the distance to the leading vehicle is substantial, and decelerate when the distance decreases below a target minimum gap. The parameters of IDM are set according to Treiber and Kesting~\cite{treiber2013traffic} as maximum acceleration $(a)=1$, maximum deceleration $(b)=1.5$, target time headway $(T)=1$, acceleration coefficient $(\delta) = 4$, and minimum gap $(s_0) = 2$.

\subsection{Heuristic-based Robot Vehicles}
FollowerStopper (FS): FS~\cite{stern2018dissipation} is an RV that travels at a fixed command velocity (target) under safe conditions but when required, slightly lowers the target velocity to open a gap to the leader vehicle. 
This allows the RV to dampen oscillations and brake smoothly when needed.  The command velocity is given by 

$$ 
v_{cmd} =
    \begin{cases}
    0, & \text{if } \Delta x \leq \Delta x_1 \\
    v \frac{\Delta x - \Delta x_1}{\Delta x_2 - \Delta x_1}, & \text{if } \Delta x_1 < \Delta x \leq \Delta x_2 \\
    v + (U - v) \frac{\Delta x - \Delta x_2}{\Delta x_3 - \Delta x_2}, & \text{if } \Delta x_2 < \Delta x \leq \Delta x_3 \\
    U, & \text{if } \Delta x_3 < \Delta x
    \end{cases}
    \label{eq:follower-stopper}
$$

\noindent where $v = \min\left(\max\left(v_{\text{lead}}, 0\right), U\right)$ is the speed of the leader vehicle, $\Delta x$ is the headway of the RV, and $U$ is the desired velocity.
The thresholds ($\Delta x_1,~\Delta x_2,~\Delta x_3$) are defined as
$$
    \Delta x_k = \Delta x_k^{0} + \frac{1}{2d_k}(\Delta v_{-})^2, \quad k = 1, 2, 3. 
    \label{eq:follower-stopper-boundary}
$$
The model parameters $\Delta x_k^{0}$, $\Delta v_{-}$, and $d_k$ together determine the spacing between vehicles and the RV's responsiveness to changes in velocity.

Proportional-integral with saturation (PIwS): PIwS~\cite{stern2018dissipation} estimates the desired average velocity ($U$) of the vehicles in the network using its historical average velocity. The PIwS RV calculates the target velocity as
$$
    v_{target} = U + v_{catch} \times \min \left( \max \left( \frac{\Delta x - g_l}{g_u-g_l}, 0 \right), 1 \right), 
    \label{eq:piws_1}
$$
which is used to calculate the command velocity at $t+1$  
$$
v_{cmd}^{t+1} = \beta_{t} (\alpha_{t} v_{target}^{t} + (1 - \alpha_{t}) v_{lead}^{t}) + (1 - \beta_{t}) v_{cmd}^{t}, 
$$ 
where $v_{catch}$ is the catch-up velocity---a velocity higher than the average velocity allows the RV to catch up with its leader, $\Delta x$ is the difference in position between the RV and its leader, $g_l$ and $g_u$ represent the lower and upper threshold distance, respectively; $\alpha_{t}$ and $\beta_{t}$ represent the weight factors for target velocity $v_{target}$ and command velocity $v_{cmd}$, respectively. Finally, $v_{lead}$ represents leader vehicle velocity.

\subsection{Model-based Robot Vehicles}
Bilateral Control Module (BCM): BCM~\cite{horn2013suppressing} uses information about both follower and leader vehicles to obtain a linear model whose acceleration is given by: 
\vspace{-.5pt}
$$
    a = k_d\cdot \Delta_d + k_v \cdot(\Delta v_l - \Delta v_f) + k_c \cdot (v_{des} - v), 
    \label{eq:bcm}
$$
where $\Delta_d$, $\Delta v_l$, $\Delta v_f$, $v_{des}$, and $v$, represent 
the difference in distance to the leader compared to the distance to the follower,
the difference in velocity to the leader, the difference in velocity to the follower, the set desired velocity, and the current velocity of the vehicle, respectively. $k_d=1$, $k_v=1$, and $k_c=1$ are gain parameters. 

Linear Adaptive Cruise Control (LACC): LACC is an improvement over existing cruise control systems that allows vehicles to maintain a safe distance or speed without communication. One implementation is the constant time-headway model by Rajamani~\cite{rajamani2011vehicle}, which employs a first-order differential equation for approximation. 
The control acceleration at time $t$ is given by
\vspace{-2pt}
$$
    a_{t}= (1 - \frac{\Delta t}{\tau})\cdot a_{(t-1)} + \frac{\Delta t}{\tau}\cdot a_{cmd, (t-1)},
$$
\vspace{-12pt}
$$
    a_{cmd}= k_1\cdot e_x + k_2 \cdot \Delta v_l,~\text{and}~e_x = s - h\cdot v,
$$
    
where $k_1=0.3$ and $k_2=0.4$ are design parameters, $e_x$ is the gap error, $s$ is the space headway, $\Delta v_l$ is the relative velocity difference to the leader, $h=1$ is the desired time gap, $\Delta t$ is the control time-step, and $\tau=0.1$ is the time lag of the control system.

\begin{figure}[t!]
    \centering
    \vspace{4pt}
    \includegraphics[width=0.90\linewidth]{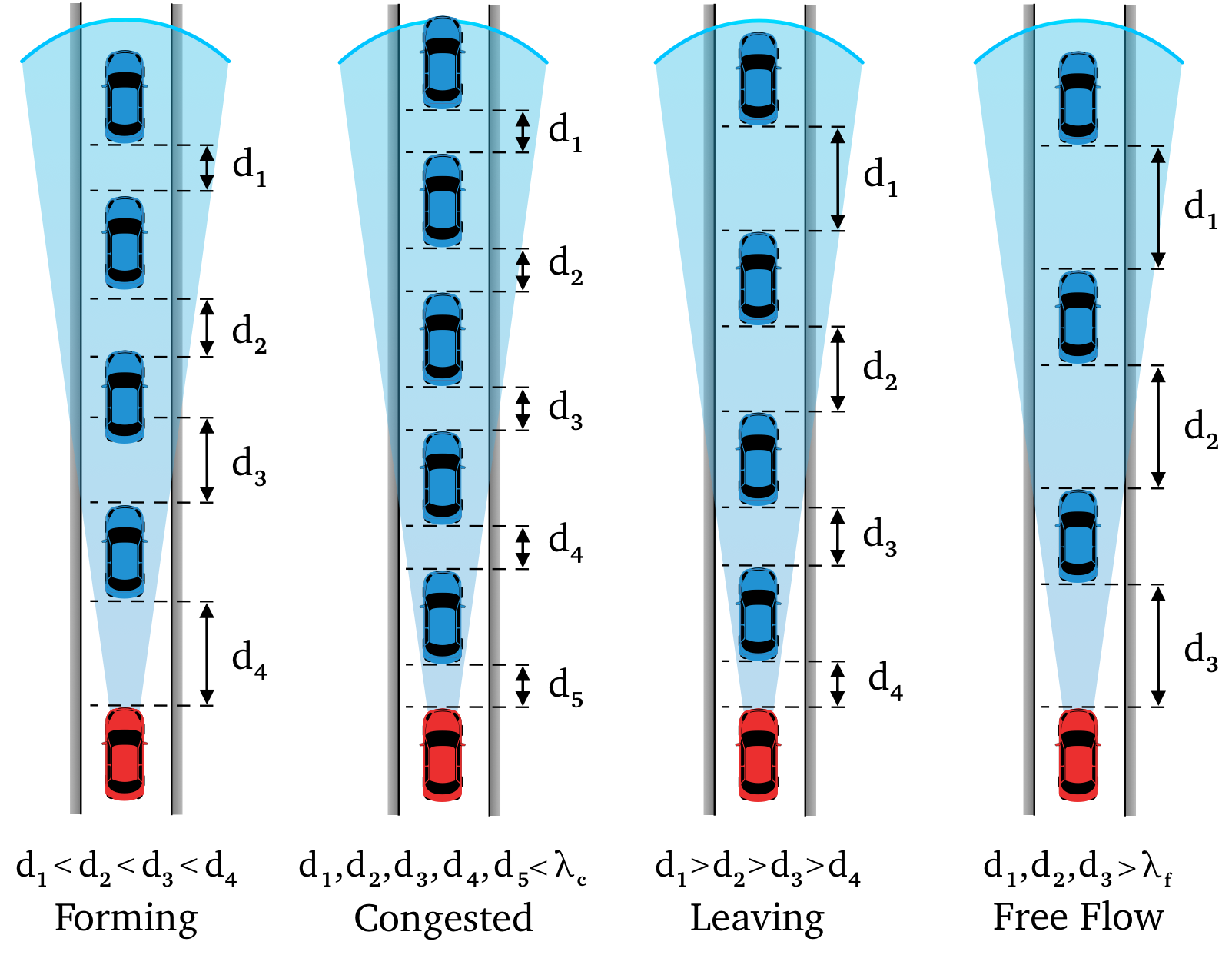}
    \vspace{-1pt}
    \caption{\small{Input data labeling for the congestion classifier (sensing zone shown in blue). The congestion classifier takes as input (position, velocity) of all vehicles in the sensing zone and outputs the traffic condition based on patterns in space headway.}} 
    \vspace{-18pt}
\label{fig:csc_labeling}
\end{figure}

\subsection{RL-based Robot Vehicles}
\label{subsec:leveraging}

RL is a $T$-step episodic task where an agent interacts with its environment to maximize the sum of discounted rewards. At each time step, the agent receives a state $s$, takes an action $a$, and the environment returns the next state $s'$ and reward $r$. This is formalized as a Partially Observable Markov Decision Process represented as $({\mathcal{S}}, {\mathcal{A}}, {\mathcal{P}}, {\mathcal{R}}, {\Omega}, {\mathcal{O}}, {\gamma})$, where ${\mathcal{S}}$ is the set of states, ${\mathcal{A}}$ is the set of actions, ${\mathcal{P}}(s', r | s, a)$ describes the environment dynamics, ${\mathcal{R}}(s,a)$ is the reward function, ${\Omega}$ is the set of observations, ${\mathcal{O}}(o | s', a)$ is the observation function, and ${\gamma}$ is the discount factor. RL-based methods have gained popularity as an effective alternative to model-based or heuristic-based methods.

Safety and efficiency goals often conflict in driving scenarios~\cite{zhu2020safe}; for example, optimizing for throughput may lead to reduced space headways and increased velocities, which can compromise safety. To address this trade-off, we propose two types of RVs: the \emph{safety} RV, which prioritizes safety, and the \emph{efficiency} RV, which emphasizes efficiency. Both RV types use the congestion classifier and operate within the same action and observation space. We train our RVs using the PPO algorithm~\cite{schulman2017proximal} with the following MDP components:

\begin{itemize}[leftmargin=*]

    \item \textbf{Observation}. The RV's observation $o_t$ at time $t$ is a combination of its own velocity ($v_t$), the relative position ($\Delta p_t$) to immediate leader HV and relative velocity ($\Delta v_t$) with respect to its immediate leader HV, and the predicted congestion stage ($c_t$) from the congestion classifier ($f_{\text{CC}}$):
    \begin{align*}
        \vspace{-5pt}
        o_t &= [v_t, \Delta p_t, \Delta v_t] \oplus c_t, \\
        c_t &= f_{\text{CC}}(\{r_{p,i}, r_{v,i}\}_{i \in \mathcal{Z}}),
        \vspace{-5pt}
    \end{align*}
    
    where $\mathcal{Z}$ denotes the set of $|\mathcal{Z}|=n$ vehicles within the sensing zone ($55~m$), and $r_{p,i}$ and $r_{v,i}$ are the relative position and velocity of the $i$-th vehicle, respectively. 
    
    \item \textbf{Action}.The RV's action ($a_t$) is its acceleration, bounded within $[-3, 3]$ m/s$^2$. 
    
    \item \textbf{Reward}. The reward function $R(s_t, a_t)$ is a weighted sum of the RV's velocity $v_t$ (for the \textit{efficiency} RV) or the average velocity of all vehicles $\bar{v}_t$ (for the \textit{safety} RV), an acceleration penalty, and a shaping term based on the predicted congestion stage $c_t$. 

    \begin{minipage}{0.99\linewidth}
        \vspace{-5pt}
        \begin{algorithm}[H]
        \caption*{\small{Reward Functions}} 
        \small{\texttt{\textit{efficiency} RV}}
        \begin{algorithmic}
        \State{$R(s_t, a_t) = 0.75 v_t - 2 |a_t|$}
        \State{\textbf{if} $c_t = \text{Congested} \wedge a_t > 0$ \textbf{then}}
        \State{\hspace{6pt} $R(s_t, a_t) \mathrel{{+}{=}} \min(-1, \lambda_1 |a_t|)$}
        \State{\textbf{if} $c_t = \text{Leaving} \wedge a_t < 0$ \textbf{then}}
        \State{\hspace{6pt} $R(s_t, a_t) \mathrel{{+}{=}} \lambda_2 |a_t|$}
        \end{algorithmic}
        
        \vspace{2pt}
        \small{\texttt{\textit{safety} RV}}
        \begin{algorithmic}
        \State{$R(s_t, a_t) = 0.15 \bar{v}_t - 4 |a_t|$}
        \State{\textbf{if} $c_t = \text{Forming}$ \textbf{then}}
        \State{\hspace{6pt} $R(s_t, a_t) \mathrel{{+}{=}} \min(-1, \lambda_3 |a_t|)$}
        
        \end{algorithmic}
        \end{algorithm}
        \vspace{-2pt}
    \end{minipage}

   where $\bar{v}_t = \frac{1}{n} \sum_{i=1}^{n} v_{i,t}$ is the average velocity of all $n$ vehicles at time $t$, and $\lambda_1 = -10$, $\lambda_2 = -10$, and $\lambda_3 = -5$ are empirically determined weights. 
    
    \item \textbf{Scaling laws}. Above $5\%$ penetration, our RVs form platoons with a single leader and multiple followers. For instance, at $40\%$ penetration, the platoon consists of $9$ RVs $(22 \times 0.4 = 8.8 \approx 9)$, with a leader RV trained at $5\%$ penetration and $8$ followers. The follower RVs observe the platoon's state (position and velocity of all vehicles in the platoon) and optimize the following reward: 
    \begin{equation*}
        R_{\text{follower}}(s_t, a_t) = \lambda_4 \Delta p_{t,j} + \lambda_5 \Delta v_{t,j} + \lambda_6 |a_{t,j}| + \lambda_7,
    \end{equation*}
    where $\Delta p_{t,j}$ and $\Delta v_{t,j}$ are the relative position and velocity of the $j$-th follower with respect to the leader, $a_{t,j}$ is the follower's acceleration, and $\lambda_4 = -2$, $\lambda_5 = 4$, $\lambda_6 = -4$, and $\lambda_7 = 10$ are empirically chosen weights. 
    \end{itemize}

\subsubsection{Congestion Classifier}
Our RL-based approach employs a congestion classifier, a neural network trained with supervised learning to predict the congestion stage $10$ time-steps in advance to allow for pre-emptive responses from our RV. The input features consist of the positions and velocities of preceding cars within the sensing zone which are mapped to one of six labels, each representing a distinct stage of congestion based on the asymmetric driving theory~\cite{yeo2008asymmetric}. The theory suggests human drivers underestimate the required space headway during deceleration and overestimate it during acceleration. Consequently, when congestion is forming, the available space headway decreases monotonically from one vehicle to the next as we move downstream within the sensing zone. Conversely, when congestion is dissipating, the available space headway increases monotonically. The six labels used to capture the congestion stages are: 'Forming', 'Leaving', 'Congested', 'Free flow', 'Undefined', and 'No Vehicle', a subset of the labels are shown in Fig.~\ref{fig:csc_labeling}.

\begin{figure}[t!]
    \centering
    \hfill
    \begin{minipage}{0.49\linewidth}
        \includegraphics[width=\linewidth]{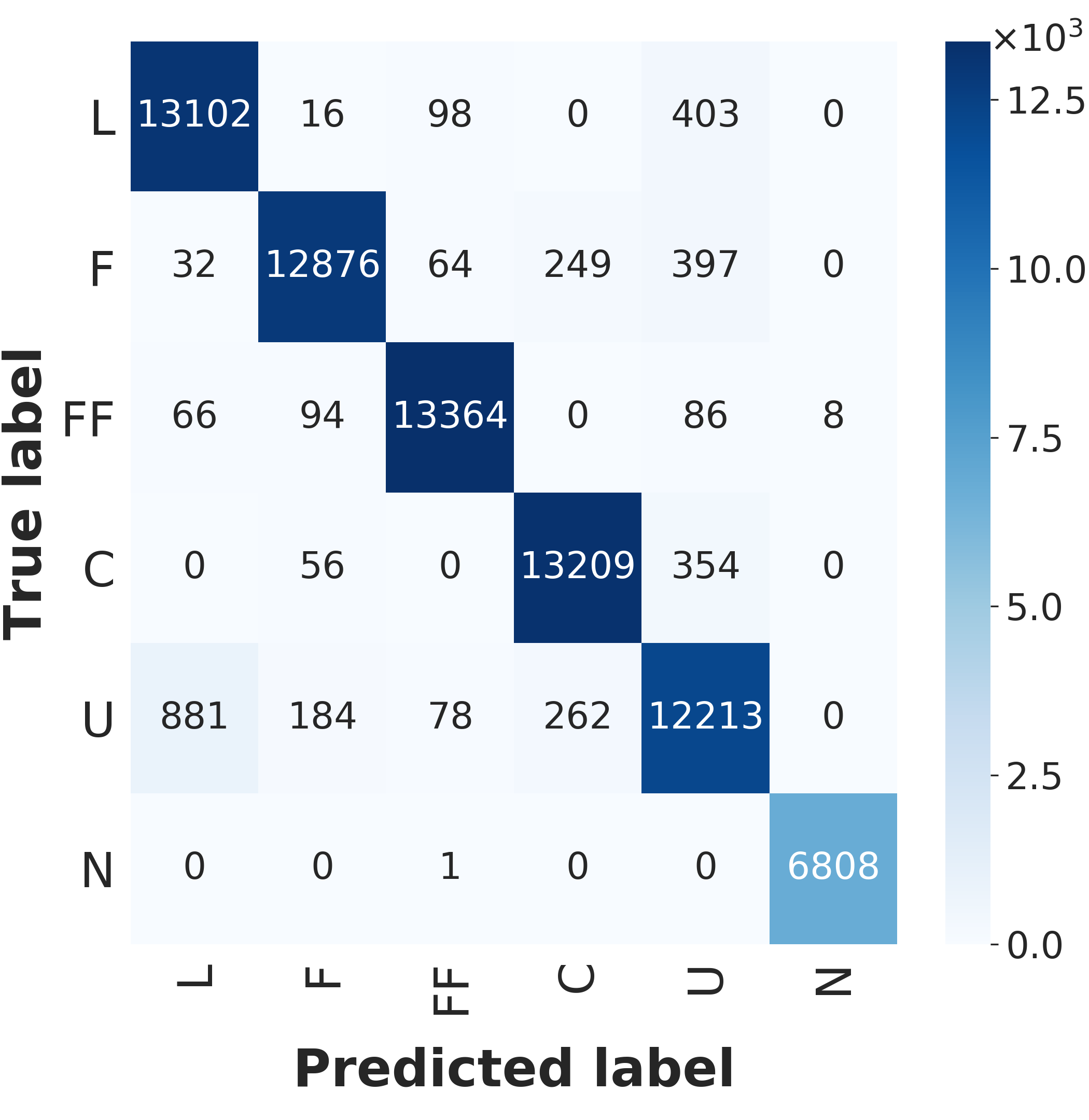}
    \end{minipage}
    \begin{minipage}{0.49\linewidth}
        \includegraphics[width=\linewidth]{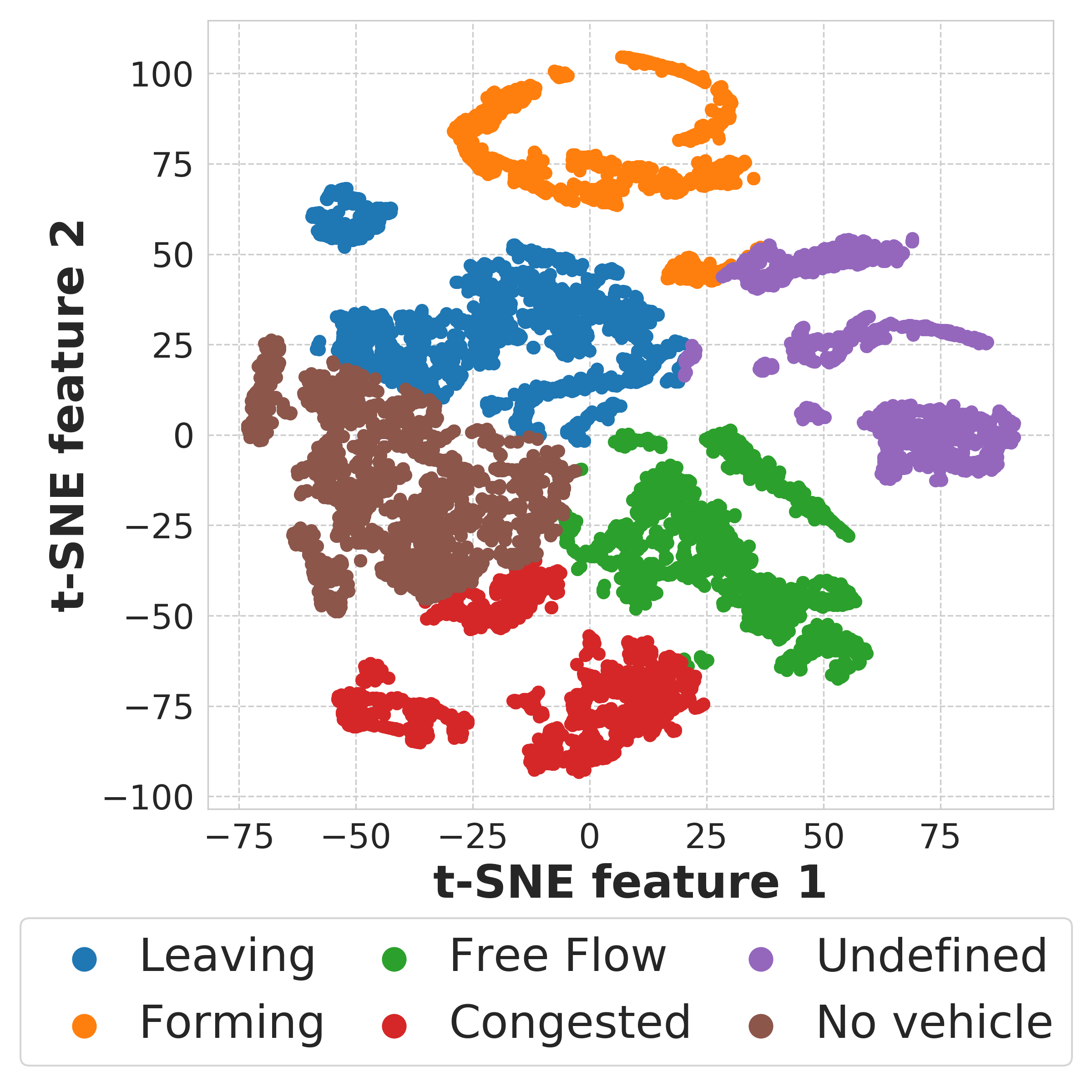}
    \end{minipage}
    \hfill
    \caption{\small{LEFT: Confusion Matrix of a trained congestion classifier in Ring on the validation set with the six classes abbreviated as: L=`Leaving', F=`Forming', FF=`Free Flow', C=`Congested', U=`Undefined', and N=`No Vehicle'. RIGHT: The results of applying K-means clustering with t-SNE on a subset of the training data of the congestion classifier. The clusters are spread out and distinct suggesting that the data is easily classifiable. }}
    \label{fig:confusion}
    \vspace{-20pt}
\end{figure}

To train the congestion classifier, we collect the position and velocity of all the vehicles inside the RV's sensing zone (set to $55~m$) at traffic density ranges $[70-133]~veh/km$. The collected data is then clustered into six classes using the K-means algorithm, as shown in Fig.\ref{fig:confusion} (RIGHT), where the dispersion of clusters indicates the classifiability of the data. Considering the sequential nature of the data and the requirement for making predictions multiple time-steps ahead, we choose a time offset of $10$ time-steps. This offset strikes a balance between the usefulness and accuracy of the predictions. A larger offset, such as $100$ time-steps, would provide more time for the RV to react, but the prediction is likely to be inaccurate, whereas a shorter offset, such as $1$ time-step, would result in accurate predictions but may not allow sufficient time for the RV to take effective action.

After windowing, the dataset includes instances where the congestion stage changes from \textit{t} to \textit{t$\plus10$}, as well as instances where the congestion stage remains the same over the time window. To train the congestion classifier, we sample data to ensure a balanced representation of transition/non-transition instances as well as instances containing all six classes. Worth noting, the `No vehicle' class presents a unique challenge. The collected data may contain instances changing from `No vehicle' to another class after the $10$ time-steps. However, based on the input corresponding to `No vehicle' at \textit{t}, we cannot predict the congestion stage at \textit{t$\plus10$}. Consequently, we discard data points where the `No Vehicle' class transitions to another class after $10$ time-steps and replace them with synthetic examples that simulate various scenarios for the RV's position and velocity without leader vehicles. The congestion classifier is trained for $50$ epochs with validation accuracy of $95.5\%$ (training parameters are provided in Table~\ref{table:params} and the confusion matrix on a validation set is shown in Fig.~\ref{fig:confusion} LEFT). Finally, We incorporate the predictions of the congestion classifier into the observations and reward function of the RV.




\subsubsection{Benchmarking RL Policies}
To benchmark with other RL techniques, we reproduce them by following the provided experiment parameters and closely matching the performance. Specifically, to obtain RL policy with only local observations, we follow Wu et al.~\cite{wu2021flow} and refer the policy as Wu hereafter; Our reproduced Wu achieves the performance within $1\%$ error (measured with stabilization time and average velocity during stabilization) of the original work.


\begin{figure*}[t!]
    \centering
    \includegraphics[width=\linewidth]{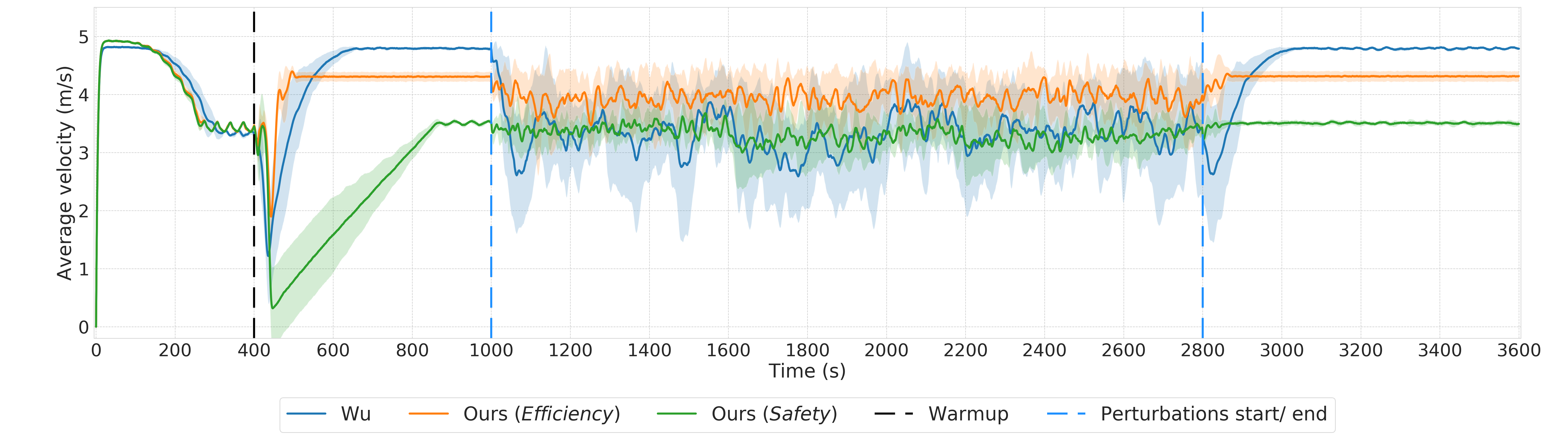}
    \vspace{-16pt}
    \caption{\small{Average velocity profile of RL-based approaches at $5\%$ penetration under long-term application of real-world perturbations (for $30$ minutes from $1000~s$ to $2800~s$), averaged over $10$ simulation rollouts. The solid lines indicate average velocity and colored ranges indicate standard deviation across rollouts. During the application of perturbations, Our \textit{efficiency} RV has the highest average velocity at $3.95~m/s$ contributing to more throughput whereas Wu has the highest standard deviation at $1.35~m/s$, indicating more sensitivity.}} 
    \vspace{-18pt}
\label{fig:speeds}
\end{figure*}

\subsection{Perturbations Via Imitation Learning and Sampling}

To ensure an accurate reproduction of real-world driving behaviors and perturbations in the simulation, we adopt a hybrid technique combining imitation learning and probabilistic sampling. We extract relevant real-world HV data during car-following such as ego velocity, headway distance, and leader velocity corresponding to the instantaneous accelerations shown in Fig.~\ref{fig:real_world} TOP. We then train a behavioral cloning model represented by an $801$-parameter neural network with these variables as input, acceleration command as predicted output, and real-world accelerations as expert demonstrations. The model minimizes the following objective:
$$
\text{MSE} = \frac{1}{N} \sum_{i=1}^{N} (a_i - \hat{a}_i)^2
$$
where $N$ is the number of samples, $a_i$ is the real-world acceleration, and $\hat{a}_i$ is the predicted acceleration. To inject acceleration perturbations to HVs in simulation, we characterize the discrete acceleration behaviors extracted from the dataset in terms of intensity, frequency, and duration. Upon analysis, we find a negative correlation of intensity with frequency and durations i.e., higher intensity accelerations tend to have shorter duration and occur less frequently. Then, we uniformly sample the frequency of the perturbations within the observed range of $[10,~30]$ per HV, for every $6$ minutes of car-following. The intensity of the acceleration perturbations is determined based on the space headway of the HV. When the space headway is less than $5~m$, human driving behavior becomes more sensitive, as the consequences of accelerations are more significant at shorter distances; hence, the accelerations are obtained from the imitation learning model to better capture the nuances of human driving. whereas if the we space headway is greater than $5~m$, we sample acceleration intensity uniformly within $[-3,~3]~m/s^2$. 
To determine the duration ($\tau_{a_i}$) of each selected intensity ${a}_i$, we first find the most common duration (mode) $\tilde{\tau}_{a_i}$ by linearly mapping $a_i$ within the minimum ($\tau{_\text{min}}$) and maximum ($\tau{_\text{max}}$) observed durations. We then sample $\tau_{a_i}$ from a piecewise triangular distribution using the following conditional probability density function:
\begin{equation*}
P(\tau_{a_i} | \tilde{\tau}_{a_i}) =
\begin{cases}
\frac{2(\tau_{a_i} - \tau_{\text{min}})}{\tau_{\text{range}}(\tilde{\tau}_{a_i} - \tau{\text{min}})}, & \tau_{\text{min}} \leq \tau_{a_i} < \tilde{\tau}_{a_i}, \\
\vspace{-10pt}\\
\frac{2(\tau_{\text{max}} - \tau_{a_i})}{\tau_{\text{range}}(\tau_{\text{max}} - \tilde{\tau}_{a_i})}, & \tilde{\tau}_{a_i} \leq \tau_{a_i} \leq \tau_{\text{max}}.
\end{cases}
\end{equation*}
where $\tau_{\text{range}} = \tau_{\text{max}}-\tau_{\text{min}}$. The sampled acceleration perturbation is randomly assigned to HVs during experiments. 

\begin{table}[h!]
\begin{center}
\vspace{8pt}
  \normalsize
  \setlength{\tabcolsep}{5pt}
  \scalebox{0.93}{
  \begin{tabular}{llr}
    \toprule
    Category & Parameter& Value \\
    \toprule    
    \multirow{3}{*}{} 
    & Time Step ($\Delta t$)& $0.1$ \\
    & Simulation Horizon ($T$) & $4500$ \\
    Simulation& Warmup Time-steps & $2500$ \\
    & Speed Limit ($m/s$) & $30$\\
    &Initial Speed ($m/s$) & $0$\\
    \hline    
    \multirow{7}{*}{PPO} 
    & Learning Rate ($\alpha$) & $0.00005$ \\
    & Discount Factor ($\gamma$) & $0.999$\\
    & GAE Estimation ($\lambda$) & $0.97$ \\
    & KL Divergence Target & $0.02$ \\
    Algorithm& Entropy Coefficient Initial& $0.1$\\
    & Entropy Coefficient Final& $0.01$\\
    & Value Function Clip Param& $20$\\
    & SGD Iterations& $2$\\
    \hline
    \multirow{3}{*}{Congestion} 
    & Neural Network& $32, 16, 16$ \\
    & Batch Size& $32$ \\
     Classifier& Learning Rate& $0.01$\\
     & Epochs& $50$\\
    \hline
    \multirow{3}{*}{Policy} 
    & Our Leader RV& $64, 32, 16$ \\
     & Our Follower RV& $64, 32, 16$ \\
     Networks& Wu& $32, 32, 32$\\ 
    \bottomrule
  \end{tabular}}
\end{center}
\vspace{-6pt}
\caption{\small{Detailed experiment parameters. We show the simulation parameters as well as the parameters of Proximal Policy Optimization (PPO) and the congestion classifier. The hidden layer dimensions of various policy networks are also shown.}}
\vspace{-18pt}
\label{table:params}
\end{table}



\section{Experiments}


\begin{table*}[t!]
\centering
\setlength{\tabcolsep}{8pt} 
\renewcommand{\arraystretch}{1.2} 
\centering
\scalebox{1.0}{
    \begin{tabular}{|c|l|cc|cc|}
    \hline
        \multirow{2}{*}{RV\%} & \multirow{2}{*}{RV Type} & \multicolumn{2}{c|}{Safety} & \multicolumn{2}{c|}{Efficiency}\\ 
        \cline{3-6} 
         & & TTC & DRAC & FE & Throughput\\ 
        \hline
          & IDM* & $1.82\thinspace\pm\thinspace0.23$ & $1.62\thinspace\pm\thinspace0.56$ & $7.63\thinspace\pm\thinspace0.23$ & $988\thinspace\pm\thinspace9.80$\\
        \cline{1-6}
        \multirow{6}{*}{\vspace{-5pt}\includegraphics[width=0.10\textwidth]{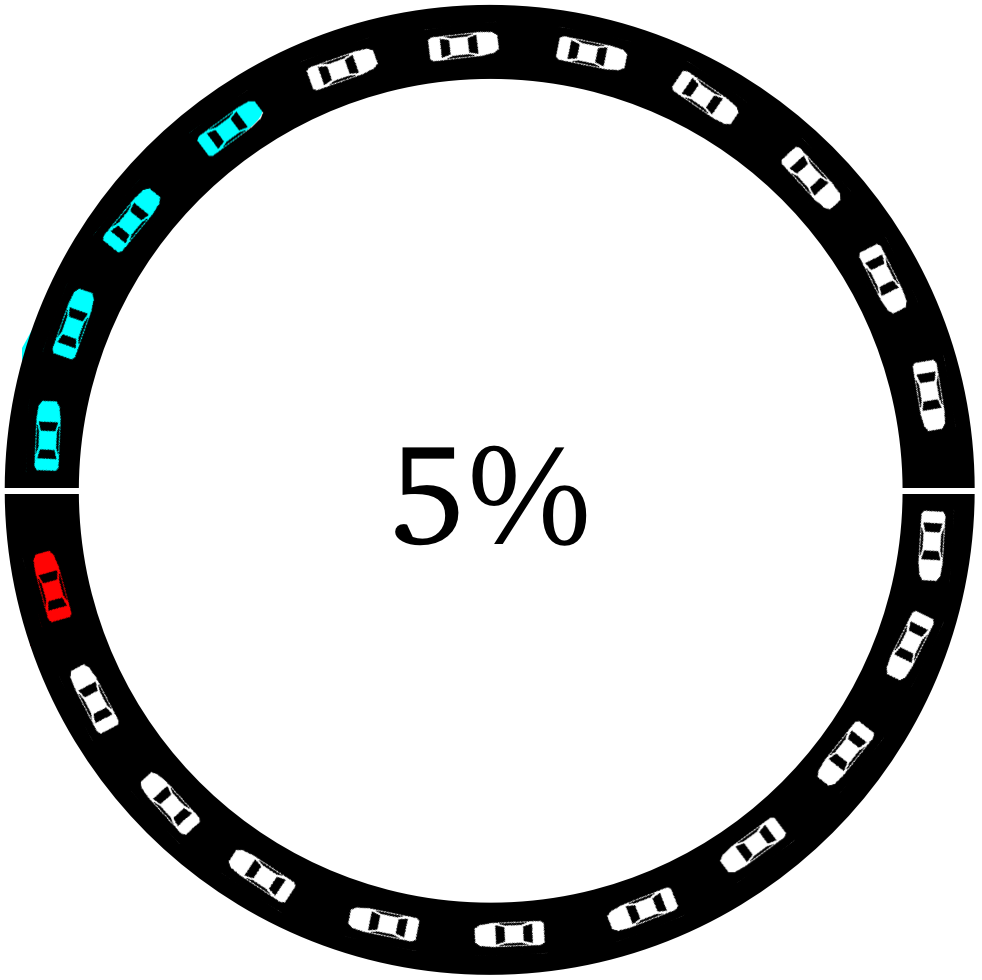}} & FS & $3.99\thinspace\pm\thinspace0.77$ & $0.89\thinspace\pm\thinspace0.27$ & $12.34\thinspace\pm\thinspace0.54$ & $1283\thinspace\pm\thinspace42.44$\\
        & PIwS & $1.95\thinspace\pm\thinspace1.57$ & $1.89\thinspace\pm\thinspace0.88$ & $12.99\thinspace\pm\thinspace0.57$ & $1343\thinspace\pm\thinspace62.62$\\
        & BCM & $1.03\thinspace\pm\thinspace0.00$ & $2.82\thinspace\pm\thinspace0.10$ & $8.17\thinspace\pm\thinspace0.19$ & $1023\thinspace\pm\thinspace14.87$\\
        & LACC & $1.09\thinspace\pm\thinspace0.01$ & $1.42\thinspace\pm\thinspace0.03$ & $8.01\thinspace\pm\thinspace0.12$ & $1031\thinspace\pm\thinspace15.13$\\
        & Wu& $2.39\thinspace\pm\thinspace0.42$ & $1.23\thinspace\pm\thinspace0.39$ & $9.92\thinspace\pm\thinspace0.66$ & $1048\thinspace\pm\thinspace82.19$\\
        & \bf{Ours} & $\bf{7.95\thinspace\pm\thinspace2.29}$ & $\bf{0.36\thinspace\pm\thinspace0.10}$ & $\bf{12.8\thinspace\pm\thinspace0.72}$ & $\bf{1242\thinspace\pm\thinspace51.34}$\\
         
        \cline{1-6}
        \multirow{6}{*}{\vspace{-5pt}\includegraphics[width=0.10\textwidth]{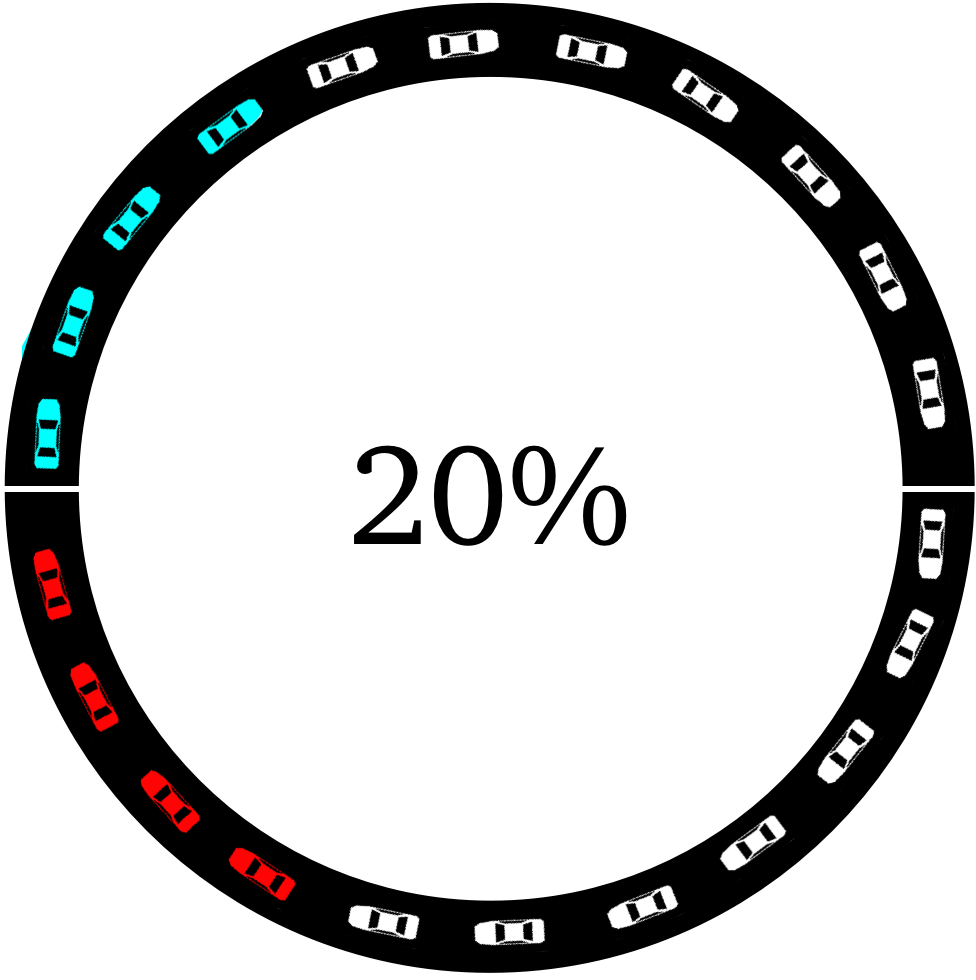}} & FS & $4.22\thinspace\pm\thinspace0.55$ & $0.74\thinspace\pm\thinspace0.26$ & $12.08\thinspace\pm\thinspace0.61$ & $1344\thinspace\pm\thinspace29.39$\\
        & PIwS & $1.71\thinspace\pm\thinspace0.18$ & $1.48\thinspace\pm\thinspace0.42$ & $11.76\thinspace\pm\thinspace0.51$ & $1328\thinspace\pm\thinspace51.73$\\
        & BCM & $2.21\thinspace\pm\thinspace1.40$ & $0.75\thinspace\pm\thinspace0.30$ & $13.19\thinspace\pm\thinspace0.36$ & $1392\thinspace\pm\thinspace18.87$\\
        & LACC & $1.08\thinspace\pm\thinspace0.04$ & $1.73\thinspace\pm\thinspace0.09$ & $9.42\thinspace\pm\thinspace0.32$ & $1166\thinspace\pm\thinspace20.59$\\
        & Wu & $2.24\thinspace\pm\thinspace0.3$ & $1.27\thinspace\pm\thinspace0.24$ & $6.41\thinspace\pm\thinspace0.17$ & $791\thinspace\pm\thinspace62.52$\\
        & \bf{Ours} & $\bf{5.84\thinspace\pm\thinspace2.96}$ & $\bf{0.59\thinspace\pm\thinspace0.28}$ & $\bf{12.87\thinspace\pm\thinspace0.64}$ & $\bf{1336\thinspace\pm\thinspace60.03}$\\ 
        \cline{1-6}
        \multirow{6}{*}{\vspace{-5pt}\includegraphics[width=0.10\textwidth]{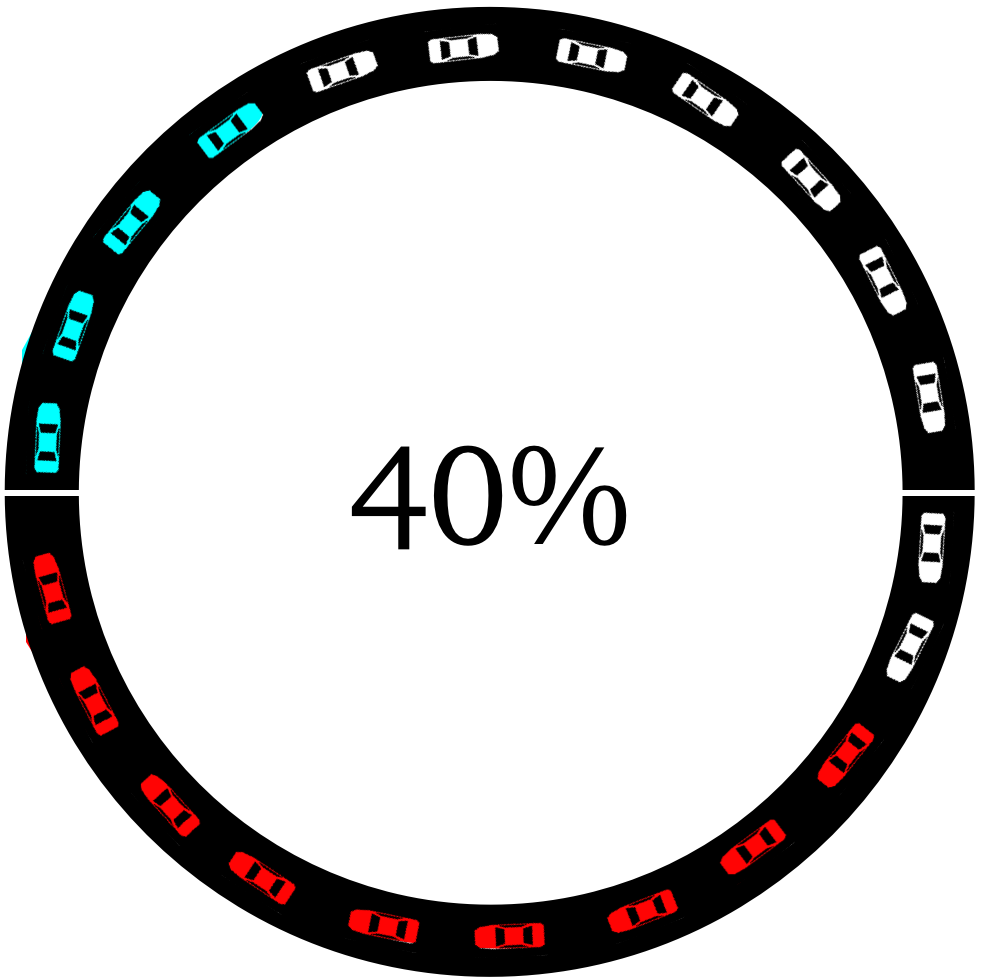}} & FS & $3.53\thinspace\pm\thinspace1.30$ & $1.02\thinspace\pm\thinspace0.60$ & $11.55\thinspace\pm\thinspace0.56$ & $1323\thinspace\pm\thinspace69.43$\\
        & PIwS & $1.58\thinspace\pm\thinspace0.16$ & $1.89\thinspace\pm\thinspace0.32$ & $11.09\thinspace\pm\thinspace0.54$ & $1294\thinspace\pm\thinspace65.91$\\
        & BCM & $4.35\thinspace\pm\thinspace6.61$ & $0.7\thinspace\pm\thinspace0.36$ & $11.96\thinspace\pm\thinspace0.42$ & $1400\thinspace\pm\thinspace10.95$\\
        & LACC & $3.14\thinspace\pm\thinspace2.34$ & $0.90\thinspace\pm\thinspace0.74$ & $14.39\thinspace\pm\thinspace0.40$ & $1426\thinspace\pm\thinspace30.07$\\
        & Wu & $2.18\thinspace\pm\thinspace0.43$ & $1.27\thinspace\pm\thinspace0.40$ & $3.66\thinspace\pm\thinspace0.16$ & $456\thinspace\pm\thinspace244.71$\\
        & \bf{Ours} & $\bf{4.22\thinspace\pm\thinspace2.08}$ & $\bf{0.89\thinspace\pm\thinspace0.52}$ & $\bf{14.06\thinspace\pm\thinspace0.45}$ & $\bf{1430\thinspace\pm\thinspace75.63}$\\
        \cline{1-6}
        \multirow{6}{*}{\vspace{-5pt}\includegraphics[width=0.10\textwidth]{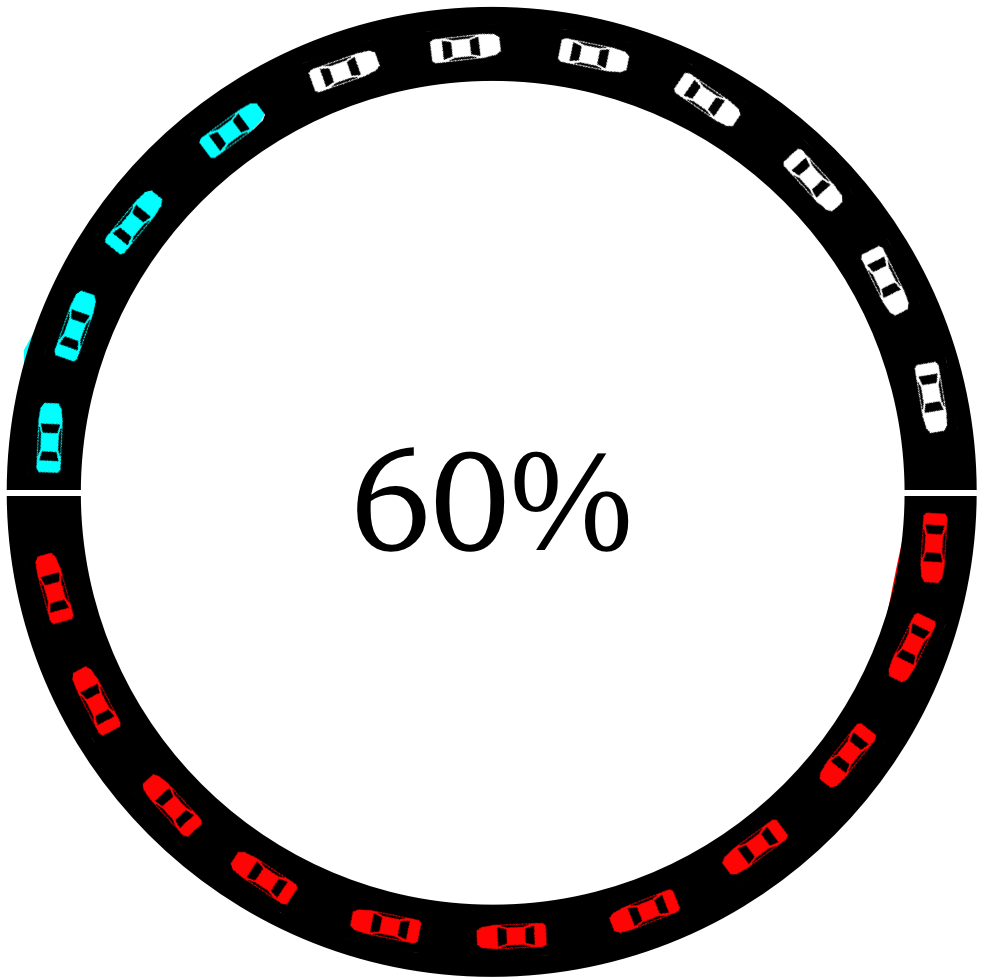}} & FS & $2.39\thinspace\pm\thinspace0.64$ & $1.00\thinspace\pm\thinspace0.05$ & $9.99\thinspace\pm\thinspace0.55$ & $1124\thinspace\pm\thinspace170.60$\\
        & PIwS & $1.59\thinspace\pm\thinspace0.30$ & $1.94\thinspace\pm\thinspace0.53$ & $10.70\thinspace\pm\thinspace0.54$ & $1264\thinspace\pm\thinspace44.99$\\
        & BCM & $2.29\thinspace\pm\thinspace1.12$ & $0.97\thinspace\pm\thinspace0.67$ & $11.79\thinspace\pm\thinspace0.49$ & $1393\thinspace\pm\thinspace6.40$\\
        & LACC & $2.55\thinspace\pm\thinspace1.26$ & $1.47\thinspace\pm\thinspace1.42$ & $13.91\thinspace\pm\thinspace0.55$ & $1448\thinspace\pm\thinspace9.80$\\
        & Wu & $2.83\thinspace\pm\thinspace1.46$ & $1.17\thinspace\pm\thinspace0.42$ & $4.41\thinspace\pm\thinspace0.10$ & $556\thinspace\pm\thinspace240.38$\\
        & \bf{Ours} & $\bf{6.58\thinspace\pm\thinspace2.47}$ & $\bf{0.57\thinspace\pm\thinspace0.27}$ & $\bf{13.61\thinspace\pm\thinspace0.54}$ & $\bf{1473\thinspace\pm\thinspace30.02}$\\
        \hline
    \end{tabular}}
\vspace{0pt}
\caption{\small{Evaluation of RVs at various penetrations (RV/s are shown in red, observed human vehicles (HVs) are shown in cyan, and remaining HVs are white) averaged over $10$ random rollouts with values after $\pm$ indicating standard deviation. IDM* denotes the $100\%$ HV baseline, Ours highlighted in bold with results of the \textit{efficiency} and \textit{safety} RVs shown Safety and Efficiency columns respectively. Across all penetrations, our \textit{safety} RV outperforms other methods in safety, exceeding the critical $4~s$ TTC threshold and reducing DRAC by up to $80\%$ compared to IDM, with the exception of $40\%$ penetration where it achieves the second-best performance in DRAC. Our \textit{efficiency} RV improves the throughput by up to $49\%$ (at $60\%$ penetration) compared to IDM, while consistently maintaining the second-highest fuel economy among all evaluated RVs, with improvement of up to $84\%$ at $40\%$ penetration.}}
\vspace{-18pt}
\label{tab:results}
\end{table*}


We introduce the mixed traffic environment, the evaluation metrics, the experimental setup, and finally the results. To begin with, we test on the mixed traffic environment, the Ring: a single-lane circular road network with $22$ vehicles as shown in Table~\ref{tab:results}. This classical scene simulates `stop-and-go traffic' where repeated cycles of accelerations and decelerations occur in HVs even in the absence of external disturbances. For our evaluations, we measure safety using surrogate measures associated with near-crash events, including Time to Collision (TTC $\uparrow$)~\cite{vogel2003comparison} to indicate collision risk and Deceleration Rate to Avoid a Crash (DRAC $\downarrow$)~\cite{cooper1976traffic} to quantify necessary braking force to avoid a collision. For efficiency, Fuel Economy (FE $\uparrow$) measures the miles driven per gallon of fuel consumed using the Handbook Emission Factors for Road Transport $3$ Euro $4$ passenger car emission model~\cite{de2004modelling}, while Throughput ($\uparrow$) indicates network capacity utilization (flow rate).

For experiment setup, we use FLOW~\cite{wu2021flow} and SUMO~\cite{lopez2018microscopic} with RVs platooned when penetration rate is $>5\%$ (all evaluated RVs can stabilize traffic in a platoon configuration, BCM and LACC require minimum $20\%$ and $40\%$ penetration respectively~\cite{chou2022lord}). We select penetration rates $5\%$, $20\%$, $40\%$, and $60\%$ to align with the minimum rates required for stabilizing traffic by different RVs~\cite{chou2022lord, poudel2023endurl}. We adhere to stringent safety standards in our evaluations of TTC, DRAC hence when multiple RVs are present, we report the worst-case values among the RVs (including  IDM baseline where the worst case among all vehicles in considered). In contrast, to measure efficiency we consider all vehicles in the network. In each experiment, we first allow the RV enough time to stabilize traffic and then we introduce acceleration perturbations for six-minute equivalent of real-world time.


Fig.~\ref{fig:speeds} shows the velocity profiles of RL-based approaches in our study, namely Wu, Our \textit{efficiency} RV, and Our \textit{safety} RV at $5\%$ penetration. All vehicles in the network are controlled using IDM during the initial $400~s$ of warmup time before the activation of RVs. After activation, Wu stabilizes traffic at $640~s$, Our \textit{efficiency} RV stabilizes traffic faster than Wu at $595~s$ whereas Our \textit{safety} RV stabilizes traffic gradually at $890~s~$. During stabilization, Wu maintains a higher average velocity of $4.88~m/s$ compared to Our \textit{efficiency} RV at $4.37~m/s$, while our \textit{safety} RV has the lowest average velocity at $3.48~m/s$. However, when perturbations are applied to HVs between $1000~s$ and $2800~s$, Our \textit{efficiency} RV maintains a higher average velocity at $3.95~m/s$ compared to Wu at $3.35~m/s$ while Our \textit{safety} RV at $3.31~m/s$ has comparable average velocity to Wu. Notably, between $1000~s$ and $2800~s$, Wu has the highest standard deviation of $1.35~m/s$ compared to Our \textit{efficiency} RV at $0.80~m/s$ and Our \textit{safety} RV at $0.70~m/s$, indicating more sensitivity to applied perturbations.


Table~\ref{tab:results} presents comparative evaluation of various RVs in the Ring at $85~veh/km$ density. Across all penetration rates, our \textit{safety} RV consistently outperforms other methods in terms of TTC. Importantly, only our RV manages to exceed the critical $4~s$ threshold for TTC at all penetration rates, a threshold that is often used to activate automatic collision avoidance systems or warn drivers~\cite{minderhoud2001extended} and has alse been recommended by earlier studies~\cite{ayres2001preferred, vogel2003comparison}. Additionally, DRAC is reduced by up to $80\%$ (at $5\%$ penetration) in comparison to IDM and Our \textit{safety} RV delivers the lowest DRAC at all penetrations except at $40\%$. Similarly, Our \textit{efficiency} RV achieves the highest throughput at higher penetration rates (improvements of $44\%$ and $49\%$ at penetration rates $40\%$ and $60\%$ respectively, compared to IDM). Furthermore, Our \textit{efficiency} RV improves fuel economy up to $84\%$ (at $60\%$ penetration) and is consistently the second highest among all other RVs at all penetration rates.

\section{Conclusion and Future Work}

In this work, we combine imitation learning and probabilistic sampling to address the \textit{Sim2Real} gap in modeling human driving behavior for mixed traffic control. Further, we propose a novel approach for optimizing safety and efficiency in mixed traffic using reinforcement learning-based robot vehicles (RVs) by introducing \model{} with two classes of RVs: \textit{safety} RV and \textit{efficiency} RV, both leveraging a classifier to predict congestion stage in advance. Through extensive experiments in the Ring with injected real-world perturbations, we demonstrate that our RV is able to increase the time to collision (TTC) above the critical $4~s$ threshold, reduce the deceleration rate to avoid a crash (DRAC) by up to $80\%$, and increase throughput up to $49\%$. \model{} is practical for real-world deployment as it relies on sensors like LiDAR which have a fixed maximum sensing range. However, additional experiments involving real hardware are needed to fully validate its scalability and robustness. 

There are also potential ethical implications regarding the safety and privacy of human drivers interacting with RVs deployed in mixed traffic that merit further investigation. \model{} inherently addresses the privacy issues as our approach prioritizes privacy by design. During the interaction between human drivers and RVs, our sensing relies on position and velocity, and does not collect sensitive information related to human drivers that could be traced back to their identity. However, a dedicated effort to investigate privacy concerns is crucial to ensure that data collected by RVs is anonymized and used responsibly.

In future work, we plan to incorporate additional traffic dynamics such as lane-changing and heterogeneous vehicle types. We also aim to conduct generalization studies in more complex environments like intersections~\cite{wang2023learning}. Additionally, we plan to perform control and coordination at a city-wide scale by relying on network-wide traffic state prediction and evaluate the robustness of such approaches~\cite{poudel2021black}. Similarly, applying our approach on real hardware, including micro-mobility vehicles~\cite{poudel2022learning}, is also an interesting future direction.

\section*{Acknowledgement}
This research is supported by NSF 2153426, 2129003, and 2038967. The authors would also like to thank NVIDIA and the Center for Transportation Research (CTR) at the University of Tennessee, Knoxville for their support. 

\bibliographystyle{unsrt}
\bibliography{references}




\end{document}